# Driver2vec: Driver Identification from Automotive Data

Jingbo Yang
jingboy@stanford.edu
Stanford University
California United States

Ruge Zhao
rugezhao@stanford.edu
Stanford University
California United States

Meixian Zhu
mxzhu@stanford.edu
Stanford University
California United States

David Hallac
david@viaduct.ai
Viaduct.ai
California United States

Jaka Sodnik
jsodnik@nervtech.com
Nervtech
Ljubljana Slovenia

Jure Leskovec
jure@cs.stanford.edu
Stanford University
California United States

## ABSTRACT

With increasing focus on privacy protection, alternative methods to identify vehicle operator without the use of biometric identifiers have gained traction for automotive data analysis. The wide variety of sensors installed on modern vehicles enable autonomous driving, reduce accidents and improve vehicle handling. On the other hand, the data these sensors collect reflect drivers' habit. Drivers' use of turn indicators, following distance, rate of acceleration, etc. can be transformed to an embedding that is representative of their behavior and identity. In this paper, we develop a deep learning architecture (*Driver2vec*) to map a short interval of driving data into an embedding space that represents the driver's behavior to assist in driver identification. We develop a custom model that leverages performance gains of temporal convolutional networks, embedding separation power of triplet loss and classification accuracy of gradient boosting decision trees. Trained on a dataset of 51 drivers provided by Nervtech, *Driver2vec* is able to accurately identify the driver from a short 10-second interval of sensor data, achieving an average pairwise driver identification accuracy of 83.1% from this 10-second interval, which is remarkably higher than performance obtained in previous studies. We then analyzed performance of *Driver2vec* to show that its performance is consistent across scenarios and that modeling choices are sound.

## 1 INTRODUCTION

Identifying the operator of a vehicle can be achieved through installing biometric devices, but there are situations in which the use of such devices is not desired [1]. In this case, a driver identification system can rely on the increasing number of sensors available on modern vehicles. Data from these sensors are already used to build applications such as automatic braking, lane departure warning, and blind spot detection [2]. These applications can be further improved through user-specific customization after inferring identity of the driver from sensor data.

A mature driver assistance system that responds to a short snippet of sensor data gives engineers the potential to build applications that adapt to specific users. At the household level, the system can quickly identify the driver behind the wheel and then adjust vehicle settings accordingly. This adjustment will not be limited to air conditioning or seat position, but will also help with vehicle maneuvering according to drivers' habits [3]. A system with high accuracy in identifying drivers can also be built to alert vehicle owners of unrecognized driving pattern to deter possibility of theft [4]. One way to build such a system is to map driving behavior to a user-specific embedding, from which many downstream tasks can also take advantage of. A primary challenge is that such system must be generalizable to all scenarios. This requirement is due to the fact that drivers behave differently according to local conditions, including different area types such as highway, urban or rural areas and road conditions like straight roads or turns. Model training for a highly capable system depends on large amounts of driving data. yet those recorded in a varieties of scenarios and by drivers of different driving styles have been difficult to obtain.

In this paper, we develop a model to map a short interval of sensor data into a driver embedding that is suitable for accurate driver identification. Our model is built to extract information from a high quality dataset and to tackle challenges outlined above. To achieve this goal, we design a customized deep learning architecture that leverages the advantages of temporal convolution, the Haar wavelet transform, triplet loss and gradient boosted decision trees' [5][6][7]. We train this model on a dataset consisting of more than 15 hours of driving data collected from a driving simulator designed by Nervtech, a high-end driving simulation company. In Section 5.1, we show that using a short 10 second snippet of driving data, we can learn embeddings that are able to identify the driver out of a set of 51 potential candidates. In a series of experiments, we further show that *Driver2vec* can correctly identify the driver in pairwise (2-way) comparison setting with 83.1% accuracy and perform consistently across different road areas. We then analyze the effects of many important hyper-parameters and model components to demonstrate the robustness of our model.

## 2 RELATED WORKS

Non-invasive identification has been investigated in many domains. For smartphone users, it has been found that movements and gestures associated with phone usage or their correlations with app usage link directly to user identity [8][9][10]. These methods show the effectiveness of using machine learning models for user identification through time series data as an alternative to facial recognition, a common invasive method for user identification.

Researchers have made direct efforts to identify vehicle operator or driving style using time series data with the following three approaches: (1) Rule-based approaches, such as fuzzy logic [11][12] (2) Supervised machine learning models such as decision trees [13], support vector machines [14][15] and neural networks [16] [17]

(3) Unsupervised/semi supervised models in the form of Gaussian mixture models (GMM) [18][19] and k-means [14][15]. Many works use a limited set of sensor data, typically only including velocity and derivatives of velocity [18] [12] [19]. Another limitation of existing works is the size of available data and the total number of participating drivers [20]. Lastly, existing methods rely on specific scenarios. One example is that machine learning models were able to achieve 76.9% accuracy only on selected turns along with discussion of difficulties of driver identification during less complex operations such as driving on straight highways [21]. Similarly, an implementation based on LSTM only used left turn maneuvers in a zone with 30km/h speed limit [17]. Our *Driver2vec* addresses all limitations discussed above. The dataset we are using has 51 unique drivers and is over 15 driving hours in total. It contains a wide variety of data channels collected from 4 driving scenarios. Furthermore, our model is trained without manual selection of desirable events or scenarios to ensure that it generalizes to all driving conditions. Since *Driver2vec* is the only work that has tackled limitations in data sources and driving scenarios, our model can serve as a cornerstone for accurate driver identification on modern vehicles.

We designed our model to take advantage of advances in similar tasks such as speaker identification via speaker embedding. In this case, performing context-free speaker diarization as an embedding task using TCN (Temporal Convolutional Network) was shown to perform as much as 6% better than alternative approaches [7]. Furthermore, the convolutional nature of a temporal convolutional network better suits hardware structure of GPUs and provides higher accuracy than traditional recurrent networks [22]. In addition to an TCN encoder, the Haar wavelet transform is also an effective method of indexing time series, often better than a discrete Fourier transform [23]. For further performance improvement, researchers have experimented with hybrid models that combine feature extraction power of neural networks with decision making power of gradient boosting decision trees, achieving 1% to 2% improvement in classification accuracy [24][25]. Despite these advances, there have been few direct applications of deep learning in vehicle sensor processing that use recurrent networks (RNN) [16][17][26]. These models face limitations such as being only able to perform well in specific driving scenarios [16] and maneuvers [17], or rely on long intervals of low frequency data [26]. Our proposed model demonstrates that driver embedding can be learned without limitations outlined above using TCN with triplet loss and that drivers can be accurately identified using LightGBM, an implementation of gradient boosting decision tree [27].

## 3 DATASET

The dataset used for this work was collected from a high-end driving simulator built by Nervtech. This simulator has been demonstrated to reproduce an environment that invokes realistic driver reactions and has been used for evaluating risks of young drivers and those with neurological diseases [28][29]. This dataset contains 51 anonymous volunteer drivers' test drives on four disjoint road areas, labeled as highway, suburban, urban and tutorial. Each driver spent approximately 15 minutes on the simulator, accumulating to more than 15 hours of driving in total. The average amount of time that drivers spend in each area is presented in Table 1. A unique advantage of this dataset is that it was designed for multiple drivers to drive the same exact scenarios on the simulator, removing external factors, including the effect of weather and road condition, from driver analysis. Another advantage of this dataset is that the simulator samples at 100Hz, much higher than datasets used by existing studies [18] [12] [19] [21] [26].

| Road Area | Average Time (SD) |
|---|---|
| Highway | 238.8 (73.8) |
| Suburban | 251.7 (83.8) |
| Urban | 196.9 (77.1) |
| Tutorial | 171.9 (83.1) |
| Total | 859.4 (245.7) |

**Table 1: Average (standard deviation) time spent in seconds per driver in each road area.**

For this study, we divided data from each road area using a 8:1:1 ratio for training, evaluation and testing. We then grouped the data columns into 8 categories: (1) distance information, (2) lane information, (3) acceleration/break pedal, (4) speed, (5) gear box, (6) acceleration, (7) steering wheel/road angle, (8) turn indicators. These categories correspond to a total of 31 features. More details of the dataset are presented in Appendix Section A.1.

## 4 DRIVER2VEC MODEL

In this paper, we convert a 10 second snippet of sensor data sampled at 100Hz to a representation which is indicative of the driver's identity. The input to our model is a multi-channeled time-series sequence. The output of our model has two stages. At first, we output an embedding of length 62 to represent a driver's behavior. Afterwards, this embedding is further processed for driver identification. To achieve this, we develop a new deep learning architecture, which we call *Driver2vec*. This new architecture is illustrated in Figure 1.

The core of *Driver2vec* architecture is time series processing. For this task, we use temporal convolutional network (TCN), which is based on dilated causal convolutions for variable length input sequences. The benefit of using a temporal convolution is that it can be scaled much larger than RNNs and often leads to significant accuracy gains in time series datasets [22]. As demonstrated by previous works, information from the frequency domain contributes to model performance [18][21]. To capture the spectral component of input channels, we use a Haar wavelet transformation, which generates two vectors in the frequency domain [30]. We pass these vectors through a linear layer then concatenate them with results from TCN as the final driver embedding vector. *Driver2vec* uses triplet loss as the training metric due to its effectiveness shown in other domains [7][31]. Each input is constructed as a set of 3 samples $\mathbf{x} = x_r, x_p, x_n$, where $x_r$ denotes an anchor, $x_p$ denotes a positive sample belonging to the same driver as $x_r$, and $x_n$ a negative sample from a different driver. The triplet loss is defined as:

$$l(x_r, x_p, x_n) = max(0, D_{rp}^2 - D_{rn}^2 + \alpha) \qquad (1)$$

where $\alpha$ is the margin and $D$ is the $l_2$ distance. In *Driver2vec*, the optimization is based on triplet loss and the objective is to achieve $D_{rn}^2 \gg D_{rp}^2 + \alpha$ using the concatenated embedding as shown in

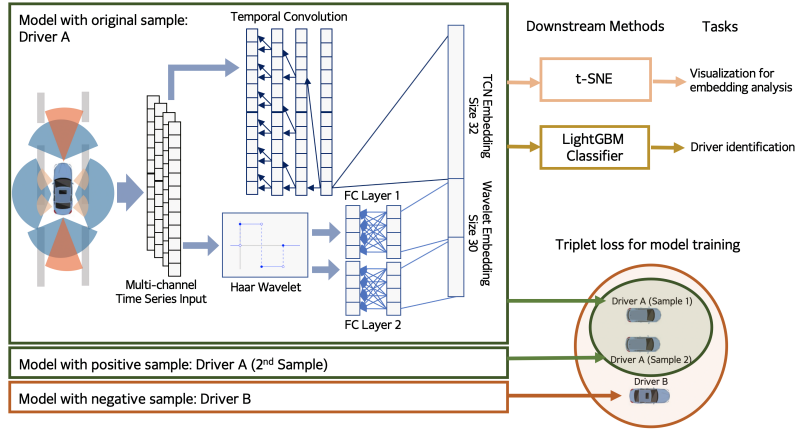

**Figure 1: Model architecture for *Driver2vec*, its loss structure and possible downstream tasks.**

Figure 1. This setup encourages embeddings for the same driver to be close to each other, and to be far apart for embedding of different drivers. Once the embedding model is trained, we use LightGBM, an efficient implementation of gradient boosting decision trees (GBDT) [6][27] for subsequent tasks. The output of the LightGBM classifier is driver identification output of the overall *Driver2vec* model, completing our time series classification workflow.

Overall, our network takes in a multi-channel time series input $\mathbb{X}$ ($\mathbb{X} \in \mathbb{R}^{31 \times 1000}$) and converts it to a 62-dimensional embedding. This embedding is then used to predict the correct driver out of potential candidates. In Section 5, we evaluate our model's performance and analyze its robustness in a series of experiments.

## 5 EVALUATION

In this section, we evaluate the performance of *Driver2vec* on a variety of driver prediction tasks. Since the *Driver2vec* embedding is trained on driver prediction, we first investigate its performance on driver identification, then analyze the importance of components of *Driver2vec*. For our experiments, we selected model based on pairwise (2-way) accuracy only. We determined that training with `Adam optimizer` using `learning rate` $= 4 \times 10^{-4}$, `decay` $= 0.975$, `kernel size` $= 16$, `triplet margin`($\alpha$) $= 1$ yields the highest pairwise evaluation accuracy of 81.8% and test accuracy of 83.1%. Hyper-parameters for the LightGBM classifier are found through grid search and are presented in Appendix Section A.2. All parameters for subsequent experiments are kept the constant unless specified otherwise.

### 5.1 Driver Prediction

We evaluate the performance of *Driver2vec* for several driver identification scenarios, which are full multi-class driver identification, identification among a set of $n$ candidates, performance in specific driving area and identification with "none-of-the-above" option.

*51-way Driver Identification.* We plot confusion matrix for driver identification result using *Driver2vec* model in Figure 2. A clear diagonal shows that the model is frequently correct. This corresponds to 15% identification accuracy among a full set of 51 candidate drivers. We then use a set of features similar to those used by a baseline approach which is based on a tree classifier for 8-10 seconds of driving data [21]. After grid searching for the best set of parameters, we are only able to achieve 10.9% identification accuracy with the baseline model. We attribute the added accuracy of our model to embeddings generated from the TCN component of *Driver2vec* because both models use frequency domain features while embeddings from TCN are unique to *Driver2vec*.

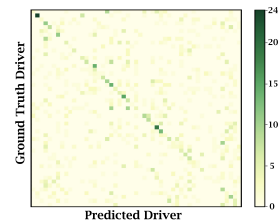

**Figure 2: Confusion matrix of 51-way driver identification.**

*Multi-way Driver Identification.* In real life, a vehicle is usually shared among a small number of drivers, therefore a model deployed in the real world should only have to identify a driver from a small group of candidates. For this purpose, we perform experiments for pairwise accuracy ($n = 2$) and $n$-way average accuracy on combinations of $n = 3, 4, 5$ drivers. Setup for $n$-way accuracy is presented in Appendix Section A.3.

Results for $n = 2, 3, 4, 5$-way accuracy are presented in Table 2. For $n = 2$, *Driver2vec* achieved 83.1% pairwise accuracy on test data. Accuracy decreased to 62.5% for 5-way identification. This decrease is in line with eventual $n = 51$ way accuracy of 15%. This observation reflects the fact that as group size increases, probability of including another driver with similar driving behavior increases, which negatively affects prediction accuracy. In comparison, we also experiment with the previously mentioned baseline [21] and are only able to achieve 69.8% for pairwise accuracy, with even wider gap in performance compared to *Driver2vec* for $n = 3, 4, 5$.

*Area Specific Prediction.* Previous work [16] demonstrated that less sophisticated roads, such as rural roads and highway, are more difficult for driver identification. However, it is important for a system deployed in the real world to perform well in all circumstances. We performed driver identification on the four areas (tutorial, urban, suburban and highway) provided by the Nervtech dataset. As shown in Table 3, the difference in test accuracy is small across different areas, at around 2%, which is much lower than 18% observed in a previous implementation [21]. This result shows that *Driver2vec* can capture much more information than velocity-related vehicle maneuver events and that environmental information, such as vehicle position in a lane and distance to next vehicle, also contribute to prediction accuracy.

| | Accuracy (Test set) (%) | Baseline (%) [21] |
|---|---|---|
| 2-way | 83.1 | 69.8 |
| 3-way | 73.8 | 55.1 |
| 4-way | 67.3 | 47.0 |
| 5-way | 62.5 | 42.7 |

**Table 2: Multi-way accuracy for *Driver2vec* and for a reference baseline implementation [21] on test data set for 2, 3, 4 and 5 way driver identification.**

| Driving Area | Pairwise Accuracy | |
|---|---|---|
| | Evaluation Set (%) | Test Set (%) |
| Highway | 78.4 | 81.1 |
| Suburban | 70.7 | 81.9 |
| Urban | 78.9 | 82.4 |
| Tutorial | 80.4 | 84.7 |

**Table 3: *Driver2vec* performance in given driving area type.**

*Prediction with None-of-the-above Option.* We introduce a "none-of-the-above" option to emulate the real world use case of identifying new or unauthorized drivers. Details for setting up the none-of-the-above evaluation is presented in Appendix Section A.3. Table 4 compares performance between with and without the none-of-the-above setup. The highest drop in performance of 15.3% is observed in the pairwise setup; the difference decreases monotonically down to 6.2% for 5-way predictions. This decrease is expected because having more candidates potentially introduces more driving styles, making it easier for the model to identify a not-included driving behavior. To our best knowledge, this work is the first to explore none-of-the-above possibility in driver identity prediction and our result shows *Driver2vec* is robust to "noise" introduced as part of our none-of-the-above experiment setup.

| | Without N-o-t-a (%) | With 50% N-o-t-a (%) |
|---|---|---|
| 2-way | 83.1 | 67.8 |
| 3-way | 73.8 | 62.2 |
| 4-way | 67.3 | 59.1 |
| 5-way | 62.5 | 56.3 |

**Table 4: Multi-way accuracy with and without none-of-the-above (N-o-t-a) noise.**

## 5.2 Model Component Analysis

In this section, we evaluate the importance of groups of sensor data and contribution of each part of *Driver2vec* to overall model performance. Analysis of two important hyper-parameters, interval length and embedding size are presented in Appendix Section A.4.

*Feature Importance.* We group similar variables in the Nervtech dataset to evaluate the importance of each category of variables. Since a model could infer velocity in $y$-direction from total velocity and its $x,z$ components, the entire group of velocity related variables must be removed together. A detailed table for grouping of these time series signals is presented in Appendix Section A.1.

| Removed Sensor Group | Pairwise Accuracy (%) |
|---|---|
| Speed, acceleration only | 66.3 |
| Distance information | 74.6 |
| Lane information | 77.8 |
| Acceleration/break pedal | 78.1 |
| Speed | 78.8 |
| Gear box | 79.0 |
| Acceleration | 79.1 |
| Steering wheel/road angle | 79.2 |
| Turn indicators | 79.3 |
| All included | **81.8** |

**Table 5: Feature ablation study through comparing pairwise-way accuracy after removal of different sensor groups.**

Table 5 shows performance of *Driver2vec* after having each group of variables removed from input. Since having "speed, acceleration only" represent a 15.5% drop in pairwise accuracy and that removing "lane information" and "distance information" also cause significant accuracy loss, drivers' behavior is much more than acceleration and turns. In fact, factors such as "lane information" and "distance information" are more important than velocity-related items, as they lead to more decrease in prediction accuracy (4% to 7.2% compared to only 1.5% to 3.7% decrease). This result implies that it is insufficient for models to only use smartphone data such as accelerometer and GPS for the purpose of drivers evaluation[32]. We further hypothesize that features groups such as "gear box", "acceleration/break pedal", "speed" and "acceleration" are highly correlated, thus our model was able to recover missing information despite having a group of features removed.

*Model Ablation.* We conduct a model ablation study to investigate the marginal benefit of each model component to overall performance. Our reference deep learning model is RNN (LSTM) with its last output connected to a linear layer for 51-way driver classification using cross-entropy loss. The next improvement is replacing LSTM with TCN for handling time series data. We then replace the cross-entropy loss function with the triplet loss setup and add wavelet features into the embeddings as illustrated in Figure 1. Performances of these incremental steps are presented in Table 6. Clearly, the most important component in *Driver2vec* is temporal convolution, supporting previous reports [22] on benefits of using TCN for time series tasks.

| Model | Pairwise Accuracy | |
|---|---|---|
| | Evaluation Set (%) | Test Set (%) |
| Baseline [21] | 67.8 | 69.8 |
| RNN Cross Entropy | 72.3 | 74.0 |
| TCN Cross Entropy | 80.4 | 80.4 |
| **TCN Triplet with LGBM** | **81.8** | **83.1** |

**Table 6: *Driver2vec* performance with different combination of components.**

## 6 A CASE STUDY: DRIVER EMBEDDING VISUALIZATION

To investigate the relationship between drivers, we plot driver embeddings, which are the inputs for the LightGBM driver classifier,

onto a 2-dimensional latent space using t-SNE. As shown on the left of Figure 3, embeddings of 3 randomly selected drivers are shown in different colors along with embeddings for rest of the drivers in gray. Although scattered, each driver's embeddings are only present in a region of the graph and have little overlap with embeddings from other drivers. In contrast, right hand side of Figure 3 contains the same background, but embeddings from a group of "indistinguishable" (for which *Driver2vec* made mistakes) drivers are colored. *Driver2vec* made incorrect predictions because these embeddings cluster in one region of the plot, indicating that they might have similar driving style.

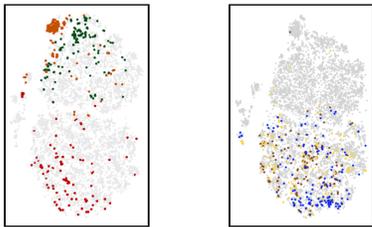

**Figure 3: Left: t-SNE visualization of 3 randomly selected drivers. Right: t-SNE visualization of 3 drivers *Driver2vec* had difficulty differentiating. (For both figures, all other drivers are colored in light gray).**

## 7   CONCLUSION

This paper proposes a deep learning architecture to map a 10 seconds snippet of sensor data into a embedding representing the driver's identity. Our proposed model, *Driver2vec*, is the first to combine temporal convolutional network with triplet loss for embedding generation. Performance on the core task of pairwise driver identification is 83.1%, beating an RNN implementation performance of 72.3% and achieves comparable accuracy against an autoencoder implementation that utilized much longer data snippets [26]. With high quality embeddings, we graphically demonstrate that embeddings from the same driver form cohesive clusters. We believe semi-supervised methods such as contrastive learning and introduction of driver stereotypes like Universal Background Model [33] for speaker identification are viable ways to improve model performance while reducing requirement on size of labeled data. To summarize, *Driver2vec* is an accurate model for driver identification. The embedding *Driver2vec* generates has great potential for many downstream applications.

## A APPENDIX

### A.1 Signals Used as Time Series Input

We grouped available data from the Nervtech dataset into the following categories, presented in Table 7. As described in Section 5.2, we removed these groups one at a time to evaluate the impact of each group on performance of *Driver2vec*. A sample of the dataset has been made publicly available on Github (repository URL has been anonymized). Source code for this study will be released after full publication.

| Group: Acceleration | |
|---|---|
| Acceleration (x) | Acceleration (y) |
| Acceleration (z) | |
| **Group: Distance information** | |
| Distance to next vehicle | Distance to next intersection |
| Distance to next stop sign | Distance to next traffic signal |
| Distance to next yield sign | Distance to completion |
| **Group: Gearbox** | |
| Gear | Clutch pedal |
| **Group: Lane information** | |
| Number of lanes present | Fast lane |
| Location in lane (right) | Location in lane (center) |
| Location in lane (left) | Lane width |
| **Group: Pedals** | |
| Acceleration pedal | Brake pedal |
| **Group: Road Angle** | |
| Steering wheel angle | Curve radius |
| Road angle | |
| **Group: Speed** | |
| Speed (x) | Speed (y) |
| Speed (z) | Speed (next vehicle) |
| Speed limit | |
| **Group: Turn indicators** | |
| Turn indicators | Tun indicators on intersection |
| **Group: Uncategorized** | |
| Horn | Vehicle heading |

**Table 7: Groups of time series signals used for *Driver2vec***

### A.2 LightGBM Hyper-parameter

Hyper-parameters for LightGBM are identified through grid search over common tuning parameters for LightGBM. The best set of hyper-parameters is presented in Table 8.

### A.3 Setup for N-way Accuracy

*N-way Accuracy.* The setup for $n = 2, 3$ is that for any given driver, we enumerated all possible subsets of size $n$ (a total of $\binom{50}{n-1}$ subsets that contain the unmasked ground truth) to generate masks. For $n = 4, 5$, we randomly sampled 4000 subsets from all possible combinations. This is because storing all $\binom{50}{3}$ subsets for 3 drivers or $\binom{50}{4}$ subsets for 4 drivers would consume excessively large amount of memory.

| LightGBM Parameter Name | Value |
|---|---|
| num leaves | 31 |
| num trees | 100 |
| max depth | 12 |
| metric | multi_logloss |
| feature fraction | 0.8 |
| bagging fraction | 0.9 |

**Table 8: Hyper-parameters for LightGBM classifier.**

*Setup for None-of-the-above.* We designed the none-of-the-above option to better represent the scenario in which an unknown driver is operating the vehicle. For example, the unknown driver can be a friend who is temporarily using the vehicle, or an indication of vehicle theft. For standard $n$-way accuracy, there is a total of $\binom{51}{n}$ tuples, out of which $\binom{50}{n-1}$ tuples contain the ground truth driver and $\binom{50}{n}$ tuples that do not contain the ground truth driver. We introduce none-of-the-above by sampling $\frac{1}{2}\binom{50}{n-1}$ tuples from those that contain the ground truth driver and another $\frac{1}{2}\binom{50}{n-1}$ from those that do not contain the ground truth driver.

At prediction time, unlike standard $n$-way accuracy where the model chooses *argmax* of predicted probabilities among $n$ probabilities, we introduce a threshold representing minimum certainty for the model to make prediction. If the sum of probabilities for candidate drivers is less than the threshold, the model's prediction will be considered "uncertain". In this case, if the ground truth is indeed not in the given set of candidates, the prediction is considered correct. With this setup, 2-way accuracy turns into a selection among (ground truth driver, another driver, not the 2 given candidates).

### A.4 Analysis of Hyper-parameters

*A.4.1 Interval Length.* We define *interval length* as the length of time in number of seconds for the time series we look ahead for the model. Recall that the Nervtech dataset is collected at 100Hz, thus interval length of 10 seconds represents an input size of $1000 \times C$, where $C$ is the number of channels ($C = 31$ for our experiments). *Interval gap* is the time interval between start of a sample. With a interval gap of 2 seconds, 5 consecutive samples will share a 2 second snippet (located at the beginning of the last sample). All experiments are conducted using the same hyper-parameters as the optimal model we found, except interval length. Figure 4 illustrates interval length and interval gap of our samples.

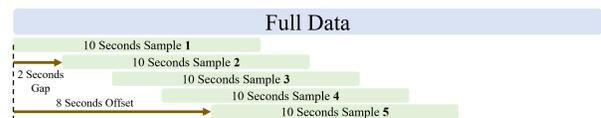

**Figure 4: Illustration of data sampling using a 10 seconds interval length with a 2 seconds interval gap.**

As illustrated in Figure 5, longer time interval leads to higher accuracy. With 30-second sequences, *Driver2vec* can achieve over 90% accuracy on evaluation data for pairwise accuracy. This observation is consistent with high driver identification accuracy achieved using stacked-RNN auto-encoder using long intervals of low frequency data [26]. However, utility of driver identification decreases as vehicle adjustments and warnings depend on responsive data processing. On the other hand, although number of samples scales inversely with interval length, model performance diminishes. This fact can be attributed to the difficulty of distinguishing between drivers of similar style in only a few seconds. This observation partially supports the slightly lower accuracy observed for highway samples, as driving on highway require less frequent adjustment of vehicle state.

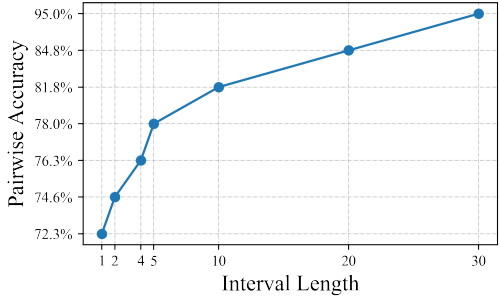

**Figure 5: Effect of interval length on pairwise accuracy with 2 seconds interval gap.**

*A.4.2 TCN Embedding Size.* Embedding size is another important hyper-parameter we investigated in detail. The size of the *Driver2vec* embedding represents the amount of information and level of complexity associated with driver behavior projected to latent space. Identifying an appropriate embedding size retains information while reduces chance for overfitting. For these experiments, we keep wavelet embedding size at 30 and only adjust the output size of TCN layers. For driver prediction, embedding from TCN and transformations from Haar wavelet are concatenated, as illustrated in Figure 1. Table 9 compares pairwise prediction accuracy with various embedding sizes. Total embedding size of 62 has the highest performance, which means a driver's behavior can be represented in approximately 62 dimensions.

| TCN embedding size (Total embedding size) | Pairwise Accuracy(%) |
|---|---|
| 16 (46) | 79.7 |
| 32 (62) | **81.8** |
| 64 (94) | 79.4 |

**Table 9: Pairwise accuracy generated on evaluation data using different embedding size.**